\documentclass[journal]{IEEEtran}
 \pdfoutput=1

\ifCLASSINFOpdf
\else
\fi
\DeclareMathAlphabet{\mathpzc}{OT1}{pzc}{m}{it}
\usepackage{hyperref}
\usepackage{url}
\usepackage{algorithm}
\usepackage{algorithmic}
\usepackage{graphicx}
\usepackage{multirow}
\usepackage{subfigure}
\graphicspath{{figs/}}
\usepackage{amsfonts}
\usepackage{color}
\usepackage{float}
\usepackage{enumerate}
\usepackage{amssymb}
\usepackage{amsmath}
\usepackage{multicol}
\usepackage{blindtext}

\def \E {\mathbb{E}}

\def\tL{\tilde{L}}

\def \xt {x_{\tau_i(t)}}

\def \g {\gamma}
\def \O {\Omega}
\def \mO {\mathcal{O}}
\def \a {\alpha}
\def \l {\lambda}
\def \r {\rho}

\def \sumn {\sum_{i=1}^n}

\def\etal{{\em et al.\/}\,}

\newtheorem{theorem}{Theorem}

\newtheorem{remark}{Remark}
\newtheorem{corollary}{Corollary}

\begin{document}
%
\title{Fast Stochastic Alternating Direction Method of Multipliers
}

\author{Leon~Wenliang~Zhong,
        James~T.~Kwok
\thanks{

Leon Wenliang Zhong and James T. Kwok are
with the Department of Computer Science and Engineering, Hong Kong
University of Science and Technology, Hong Kong.}}

\markboth{IEEE transactions, August ~2013}%
{Shell \MakeLowercase{\textit{et al.}}: Bare Demo of IEEEtran.cls for Journals}
%



\maketitle

\begin{abstract}
In this paper, we propose a new stochastic
alternating direction method of multipliers (ADMM) algorithm, which incrementally approximates the full
gradient in the linearized ADMM formulation.
Besides having a low per-iteration complexity
as existing stochastic ADMM algorithms, the proposed algorithm improves the convergence rate
on convex problems
from $\mO \left(\frac 1 {\sqrt{T}}\right)$
to $\mO
\left(\frac 1 T\right)$,
where $T$ is the number of iterations. This
matches the  convergence rate of the batch ADMM algorithm, but without the need to visit all the samples in each
iteration.
Experiments on the graph-guided fused lasso
demonstrate that the new algorithm is significantly
faster than state-of-the-art stochastic and batch
ADMM algorithms.
\end{abstract}


%
\IEEEpeerreviewmaketitle

\section{Introduction}

The alternating direction method of multipliers (ADMM) \cite{Boyd2010,Gabay1976,glowinski-75}
considers problems
of the form
\begin{equation}
\min_{x,y}\quad\Phi(x,y)\equiv \phi(x)+\psi(y) \;:\; Ax+By = c, \label{eq:obj}
\end{equation}
where $\phi, \psi$ are convex functions, and $A,B$ (resp. $c$) are
constant matrices (resp. vector) of appropriate sizes.
Because of the flexibility in splitting the objective into
$\phi(x)$ and $\psi(y)$,
it has been a popular optimization tool in many machine learning, computer vision and data
mining applications.
For example,  on large-scale distributed convex optimization, each
$\phi$ or $\psi$ can correspond to an optimization subproblem on the local data, and
the constraint $Ax+By = c$ is used to ensure all the local variables reach a global consensus;
for regularized risk minimization which will be the focus of this paper,
$\phi$ can be used  for the empirical loss,  $\psi$ for the
regularizer, and
the constraint for
encoding the sparsity pattern of the model parameter.
In comparison with other state-of-the-art optimization methods such as proximal
gradient methods \cite{Duchi2009,Xiao2010}, the use of ADMM has been shown to have faster
convergence in several
difficult structured sparse regularization problems \cite{Suzuki2013}.

Existing works on ADMM often assume that $\Phi(x,y)$ is deterministic. In the context of
regularized risk minimization, this corresponds to
batch learning and
each iteration needs to visit all the samples.
With the proliferation of data-intensive applications,
it can quickly become computationally expensive.
For example, in using ADMM on the overlapping group-lasso, the matrix computations
become costly when both the number of features and data set size are large~\cite{Qin2012}.
To alleviate this problem,
the use of
stochastic and online techniques have recently drawn a lot of interest.
Wang and Banerjee~\cite{Wang2012} first
proposed the online
ADMM, which learns from only one sample (or a small mini-batch) at a time. However, in general,
each round involves nonlinear optimization and is not computationally appealing.
Very recently, three
stochastic variants of ADMM are independently proposed
\cite{Ouyang2013,Suzuki2013}.
Two are based on the stochastic gradient descent (SGD) \cite{bottou-04},
while one is based on regularized dual averaging
(RDA)~\cite{Xiao2010}. In both cases, the difficult nonlinear optimization problem inherent in
the online ADMM is circumvented by
linearization,
which
then  allows
the resultant iterations in these stochastic variants to
be efficiently
performed.

However, despite their low per-iteration complexities, these stochastic ADMM algorithms converge at a
suboptimal rate
compared to their batch counterpart.
Specifically, the algorithms in
\cite{Ouyang2013,Suzuki2013} all achieve a rate of
$\mO(1/\sqrt{T})$,
where $T$ is the number of iterations,
for general convex problems
and
$\mO(\log T/T)$
for strongly convex problems;
whereas batch ADMM achieves convergence  rates of
$\mO(1/T)$ and
$\mO(\mu^T)$ (where $0<\mu<1$), respectively
\cite{Deng2012,He2011}.
This gap in the convergence rates between stochastic and batch ADMM algorithms is indeed not surprising, as it is also observed between
SGD and batch gradient descent
in the analogous unconstrained optimization setting
\cite{Mairal2013}.
Recently, there have been several attempts on bridging this gap
\cite{LeRoux2012,Mairal2013,Shalev-Shwartz2013}.
For example, Re Loux \etal \cite{LeRoux2012} proposed an approach whose
per-iteration cost is as low as SGD, but can achieve linear convergence
for strongly convex functions.

Along this line, we propose  in the sequel a novel stochastic algorithm that bridges the
$\mO(1/\sqrt{T})$ vs $\mO(1/T)$ gap in convergence rates for
ADMM.  The new algorithm enjoys the same computational simplicity  as existing stochastic ADMM
algorithms, but with a
much faster convergence rate matching that of its batch counterpart. Experimental results
demonstrate that it dramatically outperforms existing stochastic and batch ADMM algorithms.

{\bf Notation.}
In the sequel, we use $\|\cdot\|$ to denote a general norm, and $\|\cdot\|_*$ be its dual norm (which is
defined as $\|z\|_*=\sup_{x:\|x\|\leq 1} z^Tx$).
For a function $f$, we let
$f'$
be a subgradient
in the subdifferential set
$\partial f(x)=\{g \;|\; f(y) \geq f(x)+g^T(y-x), \; \forall y\}$.
When $f$ is differentiable, we use $\nabla f$ for its gradient.
A function $f$ is $L$-Lipschitz if $\|f(x) - f(y)\|\leq L\|x-y\| \; \forall x,y$.
It is $L$-smooth if
$\|\nabla f(x) - \nabla f(y)\|_*\leq L\|x-y\|$, or equivalently,
\begin{equation} \label{eq:smooth}
f(y) \leq f(x) + \nabla f(x)^T(y-x)+ \frac{L}{2}\|x-y\|^2.
\end{equation}
Moreover, a function $f$ is $\mu$-strongly convex if $f(y) \geq f(x) + g^T(y-x) +
\frac{\mu}{2}\|y-x\|^2$
for $g \in
\partial f(x)$.


\section{Batch and Stochastic
ADMM}
\label{sec:admm}

As in the method of multipliers,
ADMM starts with the augmented Lagrangian of problem~(\ref{eq:obj}):
\begin{equation} \label{eq:lag}
L(x,y,u) = \phi(x)+\psi(y)+\beta^T(Ax+By-c)
+\frac{\rho}{2}\|Ax+By-c\|^2,
\end{equation}
where $\beta$ is the vector of Lagrangian multipliers, and $\rho>0$ is a penalty
parameter.
At the $t$th iteration,  the values of $x$ and $y$ (denoted
$x_t,y_t$) are updated by minimizing $L(x,y,\beta)$ w.r.t.  $x$ and $y$.
However, unlike the method of multipliers,
these are minimized in an alternating manner,
which allows the problem
to be more easily decomposed
when $\phi$ and $\psi$ are separable.
Using the
scaled dual variable  $\a_t = \beta_t/\rho$,
the ADMM update can be
expressed
as \cite{Boyd2010}:
\begin{eqnarray}
x_{t+1} & \leftarrow & \arg \min_{x} \; \phi(x)+\frac {\rho} 2\|Ax+By_t-c+\a_t\|^2, \label{eq:up_x}\\
y_{t+1} & \leftarrow & \arg \min_{y} \; \psi(y)+\frac {\rho}
2\|Ax_{t+1}+By-c+\a_t\|^2, \label{eq:up_y}\\
\a_{t+1} & \leftarrow & \a_{t} + Ax_{t+1}+By_{t+1}-c. \label{eq:up_a}
\end{eqnarray}

In the context of regularized risk minimization,  $x$ denotes the model parameter to be learned. Moreover,
we assume that
\begin{equation} \label{eq:phi}
\phi(x)\equiv\frac 1 n\sum_{i=1}^n \ell_i(x)+\O(x),
\end{equation}
where $n$ is the number of samples, $\ell_i(x)$
denotes sample $i$'s contribution to the empirical loss, and $\O(x)$ is a ``simple''
regularizer commonly encountered in proximal methods. In other words, $\O$ is chosen such that the proximal step
$\min_{x} \frac{1}{2}\|x-a\|^2+\O(x)$, for some constant $a$, can be computed efficiently.
Subproblem (\ref{eq:up_x}) then becomes
\begin{equation} \label{eq:admm_x}
x_{t+1} \; \leftarrow \; \arg \min_{x} \; \frac 1 n\sum_{i=1}^n \ell_i(x) +\O(x)+\frac {\rho} 2\|Ax+By_t-c+\a_t\|^2.
\end{equation}

When the data set is large,
solving (\ref{eq:admm_x}) can be computationally expensive.
To alleviate this problem, Wang and
Banerjee~\cite{Wang2012} proposed the {\em online ADMM} that uses only
one sample in each iteration. Consider the case where $\O=0$. Let the index of the sample selected at iteration $t$ be $k(t) \in
\{1,2,\ldots,n\}$.
Instead of using (\ref{eq:admm_x}),
$x$ is updated
as
\begin{equation} \label{eq:wang_up_x}
x_{t+1}  \leftarrow  \arg \min_{x}  \;
\ell_{k(t)}(x)+\frac {\rho} 2\|Ax+By_t-c+\a_t\|^2+\frac{D(x,x_t)}{\eta_{t+1}},
\end{equation}
where $D(x,x_t)$ is a Bregman divergence between
$x$ and $x_t$,
and  $\eta_t \propto \frac{1}{\sqrt{t}}$
is the stepsize.
In general,
solving (\ref{eq:wang_up_x}) requires the use of  nonlinear optimization,
making it expensive and less appealing. Very recently,
several stochastic versions of ADMM have been independently proposed.
Ouyang \etal \cite{Ouyang2013} proposed the {\em stochastic ADMM}\footnote{To avoid confusion with other
stochastic variants of ADMM, this particular algorithm will be called STOC-ADMM in the sequel.}, which
updates $x$ as
\begin{eqnarray}
x_{t+1}  & \leftarrow  & \arg \min_{x}  \;
\ell_{k(t)}'(x_t)^T(x-x_t)
+\frac {\|x-x_t\|^2}{2\eta_{t+1}} \nonumber \\
& & +\frac {\rho} 2\|Ax+By_t-c+\a_t\|^2 \label{eq:ouyang_up_x}
\\
& = & \left(\frac 1{\eta_{t+1}}I+\r A^TA\right)^{-1} \cdot \nonumber \\
& & \left[\frac{x_t}{\eta_{t+1}}-\ell_{k(t)}'(x_t)-\r A^T( By_t-c+\a_t)\right], \label{eq:ouyang_up_x2}
\end{eqnarray}
where
$\ell_{k(t)}'(x_t)$ is the (sub)gradient of $\ell_{k(t)}$ at $x_t$,
$\eta_t\propto \frac{1}{\sqrt{t}}$ is the stepsize,
and $I$ is the identity matrix of appropriate size.
The updates for $y$ and $\a$ are the same as in (\ref{eq:up_y}) and (\ref{eq:up_a}).
For the special case where $B=-I$ and $c=0$,
Suzuki~\cite{Suzuki2013} proposed a similar approach called {\em online proximal gradient descent
ADMM} (OPG-ADMM), which uses the inexact Uzawa method  \cite{zhang-11} to further linearize the last term in
(\ref{eq:ouyang_up_x}),
leading to
\begin{eqnarray}
x_{t+1}  & \leftarrow  & \arg \min_{x} \; (\ell_{k(t)}'(x_t)+ \r A^T(Ax_t-y_t+\a_t))^Tx \nonumber \\
& & +
\frac {\|x-x_t\|^2}{2\eta_{t+1}} \label{eq:opg_up_x} \\
& = & x_t - {\eta_{t+1}}\left[ {\ell_{k(t)}'(x_t)+ \r A^T(Ax_t-y_t+\a_t)}\right].
\label{eq:suzuki}
\end{eqnarray}
Compared to
(\ref{eq:ouyang_up_x2}), it avoids the inversion of
$\frac 1{\eta_{t+1}}I+\r A^TA$
which can be computationally expensive
when $A^TA$ is large.
Suzuki~\cite{Suzuki2013} also proposed another stochastic variant called {\em RDA-ADMM}
based on the method of regularized dual averaging (RDA) \cite{Xiao2010}
(again for the
special case with $B=-I$ and $c=0$),
in which $x$ is updated as
\begin{eqnarray}
x_{t+1}  &   \leftarrow&\arg\min_x (\bar{g}_t+ \r A^T(A \bar x_t^{\text{RDA}} -\bar y_t^{\text{RDA}}
+\bar \a_t^{\text{RDA}} )^T x
\nonumber \\
& &
+\frac {\|x\|^2}{\eta_{t+1}}\label{eq:rda_up_x}\nonumber\\
 &=&-\eta_{t+1}\left[\bar{g}_t + \r A^T(A \bar x_t^{\text{RDA}} -\bar y_t^{\text{RDA}}  +\bar \a_t^{\text{RDA}}) \right]. \label{eq:rda_up_x}
 \end{eqnarray}
Here, $\eta_t\propto \sqrt{t}$, $\bar{g}_t=\frac 1 t \sum_{j=1}^t \ell_{k(j)}'(x_j), \bar
x_t^{\text{RDA}} = \frac 1 t
\sum_{j=1}^t x_j, \bar y_t^{\text{RDA}} =\frac 1 t \sum_{j=1}^t y_j$, and $\bar \a_t^{\text{RDA}}  =\frac 1 t \sum_{j=1}^t
\a_j$ are
averages obtained from the past $t$ iterations.

For general convex problems,
these online/stochastic ADMM approaches
all converge at a rate
of $\mO(1/\sqrt{T})$, w.r.t. either the objective value
\cite{Suzuki2013} or a weighted combination of the objective value and
feasibility violation \cite{Ouyang2013}.
When $\phi$ is further required to be strongly convex, the convergence rate
can be improved to $\mO(\log T/T)$
(except for RDA-ADMM whose convergence rate in this situation is not clear). However,
in both cases (general and strongly convex), these convergence rates are inferior to their batch
ADMM counterparts,
which are
$\mO(1/T)$ and $\mO(\mu^{T})$ (with $0<\mu<1$),
respectively
\cite{Deng2012,He2011,hong-12}.



\section{Stochastic Average ADMM (SA-ADMM)}

On comparing the update rules on $x$ for the STOC-ADMM and OPG-ADMM
((\ref{eq:ouyang_up_x}) and (\ref{eq:opg_up_x})) with that of the
batch ADMM (\ref{eq:admm_x}), one can  see that
the empirical loss on the whole training set (namely, $\frac 1 n\sum_{i=1}^n
\ell_i(x)$)
is replaced
by the linear approximation based on one single sample plus a proximal term $\frac
{\|x-x_t\|^2}{2\eta_{t+1}}$.  This follows the standard approach taken by
SGD. However, as is well-known, while the full gradient
descent has linear
convergence rate,
SGD only achieves sublinear convergence \cite{Mairal2013}.
This agrees with the result
in Section~\ref{sec:admm} that the existing stochastic versions of ADMM all have
slower convergence rates than their batch counterpart.

Recently, by observing that the training set is indeed finite,  it is shown that
the convergence rates of stochastic algorithms can be improved to match those of the batch learning algorithms.
A pioneering approach along this line is the {\em stochastic average gradient}
(SAG)~\cite{LeRoux2012}, which considers the optimization of a strongly convex sum of smooth functions
($\min_x\frac 1 n\sum_{i=1}^n \ell_i(x)$).
By updating an estimate of the full gradient incrementally
in each iteration,
its per-iteration time complexity is only as low as SGD,
yet surprisingly its convergence rate is linear.

Another closely related approach is
the {\em minimization by incremental surrogate optimization} (MISO)
\cite{Mairal2013}, which replaces each $\ell_i$ by some ``surrogate'' function in an incremental
manner similar to SAG. In particular, when each $\ell_i$ is smooth and the so-called ``Lipschitz
gradient surrogate" is used,
the resultant update rule is very similar to that of SAG.

Motivated by these recent stochastic optimization results,
we will propose in the following a novel stochastic ADMM algorithm
that achieves the same convergence rate as batch ADMM on
general convex
problems.
Unlike SAG or MISO, the proposed algorithm is more general and can be applied to optimization problems with equality
constraints, which are naturally handled by ADMM.


\subsection{Algorithm}

In the following, we assume that  the $\ell_i$ in (\ref{eq:phi}) is $L$-smooth (e.g., the square loss and logistic loss).
As in existing stochastic ADMM approaches \cite{Wang2012, Ouyang2013, Suzuki2013},
$y$ and $\a$
are still updated by (\ref{eq:up_y}), (\ref{eq:up_a}), and the key difference is on the update of
$x$
(Algorithm~\ref{alg:sadmm}).
First,
consider the special case where
$\O =0$.
At iteration $t$, we randomly choose a sample $k(t)$
uniformly
from $\{1,2,\dots,n\}$, and
then update
$x$ as
\begin{equation} \label{eq:our_up_x1}
x_{t+1}\leftarrow \arg\min_{x}
 P_t^{\ell}(x)
+r(x,y_t,\a_t),
\end{equation}
where $P_t^{\ell}(x)\equiv\frac{1}{n}\sumn \ell_i(\xt)+\nabla \ell_i(\xt)^T(x-\xt)+ \frac {L} {2} \|x-\xt\|^2$, $r(x,y,\a)\equiv\frac {\r}2\|Ax+By-c-\a\|^2$,
and
$\tau_i(t)= \left\{ \begin{array}{ll} t & i=k(t) \\ \tau_i(t-1) & \text{otherwise}
\end{array} \right.$.
Hence, as in SAG, out of the $n$ gradient terms in (\ref{eq:our_up_x1}),
only one of them (which corresponds to sample $k(t)$)
is based on the current iterate $x_t$, while all
others are previously-stored gradient values. Moreover,
note its similarity with the STOC-ADMM update in (\ref{eq:ouyang_up_x}), which only retains terms related to
$\ell_{k(t)}$,
while (\ref{eq:our_up_x1}) uses the information from all of $\{\ell_1,\ell_2,\ldots,\ell_n\}$.
Another difference with
(\ref{eq:ouyang_up_x}) is that the proximal term in (\ref{eq:our_up_x1}) involves a constant
$L$, while
(\ref{eq:ouyang_up_x2}) requires a time-varying stepsize $\eta_t$.

By setting the derivative of (\ref{eq:our_up_x1}) to zero, we have
\begin{equation} \label{eq:our_upd_x}
x_{t+1}\leftarrow (\r A^TA+LI)^{-1} \left[L\bar x_t-\r A^T(By_t-c+\a_t) - \overline{\nabla \ell}_t\right],
\end{equation}
where $\bar x_t \equiv \frac 1 n\sum_{i=1}^n x_{\tau_i(t)}$, and
$\overline{\nabla \ell}_t \equiv \frac 1 n\sum_{i=1}^n \nabla \ell_i(x_{\tau_i(t)})$.
When the dimension of $A^TA$ is manageable,
(\ref{eq:our_upd_x}) is cheaper than the STOC-ADMM update,
as
$(\r A^TA+LI)^{-1}$
can be pre-computed and stored;
whereas
$\eta_t$
in (\ref{eq:ouyang_up_x2})
is changing
and consequently  the matrix inverse there
has to be re-computed
in every iteration.\footnote{Ouyang \etal \cite{Ouyang2013} claimed that $\eta_{t+1}$
in (\ref{eq:ouyang_up_x2}) can be fixed at $\eta_T$, though
the proof is missing. Moreover, as $\eta_t$ is decreasing, $\eta_T$ is the most conservative
(smallest) stepsize. Hence, using
$\eta_T$ in all iterations
may lead to very slow
convergence
in practice.}
When $\r A^TA+LI$ is large, even storing its inverse can be
expensive.
A technique that has been popularly used in the recent ADMM  literature is the inexact
Uzawa method \cite{zhang-11}, which
uses (\ref{eq:smooth}) to
approximate
$r(x,y_t,\a_t)$ by its
upper bound:
\begin{eqnarray}
r(x,y_t,\a_t) & \leq & P_t^{r}(x) \nonumber \\
& \equiv & r(x_t,y_t,\a_t)+\nabla_{x}^t r^T(x-x_t) \nonumber \\
& & +\frac {L_{A}}2\|x-x_t\|^2,
\label{eq:r}
\end{eqnarray}
where $\nabla_{x}^t r\equiv \r A^T(Ax_t+By_t-c+\a_t)$ and
$L_A$ is an upper bound on the eigenvalues of $\r A^TA$
\cite{He2011}.
Hence, (\ref{eq:our_up_x1}) becomes
\begin{eqnarray}
x_{t+1}&\leftarrow& \arg\min_{x}P_t^{\ell}(x)+P_t^{r}(x) \label{eq:uzawa1} \\
&=&\frac {L\bar x_t +L_A x_t -\left[\overline{\nabla \ell}_t+\nabla_{x}^t r)\right]}{L_A+L}.
\label{eq:uzawa}
\end{eqnarray}
Analogous to the discussion between (\ref{eq:our_up_x1})
and (\ref{eq:ouyang_up_x}) above,
our (\ref{eq:uzawa})
is also similar to the OPG-ADMM update in (\ref{eq:suzuki}),  except that all the
information from $\{\ell_1,\ell_2,\ldots,\ell_n\}$ are now used. Moreover, note that although
RDA-ADMM
also uses an average of gradients
($\bar{g}_t$ in (\ref{eq:rda_up_x})), its convergence
is still slower than the proposed algorithm (as will be seen in Section~\ref{sec:conv}).

When
$\O\neq 0$, it can be added back to
(\ref{eq:our_up_x1}),  leading to
\begin{equation} \label{eq:omega}
x_{t+1}\leftarrow \arg\min_{x} P_t^{\ell}(x) +\O(x) +r(x,y_t,\a_t).
\end{equation}
In general, it is easier to solve with the inexact Uzawa simplification. The update then
becomes
\begin{eqnarray}
\!\!\! x_{t+1}&\!\!\! \leftarrow\!\!\!& \arg\min_{x}P_t^{\ell}(x)+P_t^{r}(x)+\O(x) \label{eq:iu}\\
\!\!\! &\!\!\! =\!\!\! &\arg\min_x \frac 1 2\left \|x-\frac {L\bar x_t +L_A x_t -\left[\overline{\nabla
\ell}_t+\nabla_{x}^t r\right]}{L_A+L}\right\|^2 \nonumber\\
 &\!\!\! + \!& \frac{\O(x)}{L_A+L}.
\label{eq:our_up_x2}
\end{eqnarray}
This is the standard proximal step popularly used in optimization problems with structured
sparsity \cite{Bach2011,Beck2009}.
As is well-known, it can be efficiently computed
as $\O$ is assumed ``simple'' (e.g., $\O(x)=\|x\|_1$, $\|x\|_2$, $\|x\|_{\infty}$ and various mixed norms \cite{Duchi2009}).

\begin{algorithm}[t]
\caption{Stochastic average alternating direction method of multipliers (SA-ADMM).}
\label{alg:sadmm}
\begin{algorithmic}[1]
\STATE {\bfseries Initialize:} $x_0,y_0,\a_0$ and $\tau_i(-1)=0 \forall i$.
\FOR{$t=0,1,\ldots,T-1$}
\STATE randomly choose $k\in \{1,2,\dots, n\}$, and set $\tau_i(t)=\begin{cases}
  t \qquad \quad\text{ if }i=k\\
  \tau_i(t-1) \text{ otherwise}
\end{cases}$;
\STATE update $x_{t+1}$ using (\ref{eq:our_upd_x}) or (\ref{eq:uzawa}) when $\O=0$; and
use (\ref{eq:our_up_x2}) when
$\O\neq 0$;
\STATE $y_{t+1}\leftarrow \arg\min_{y} \psi(y)+\frac {\r}2\|Ax_{t+1}+By-c+\a_t\|^2$; \label{alg:sadmm_upy}
\STATE $\a_{t+1}\leftarrow \a_t+(Ax_{t+1}+By_{t+1}-c)$; \label{alg:sadmm_upa}
\ENDFOR
\STATE {\bfseries Output:} $\bar x_T\leftarrow \frac 1 T\sum_{t=1}^T x_t,\bar y_T\leftarrow\frac 1 T\sum_{t=1}^T y_t$.
\end{algorithmic}
\end{algorithm}


\subsection{Discussion}

In the special case where $\psi=0$, (\ref{eq:obj}) reduces to $\min_x \phi(x)$ and the
feasibility violation
$r(x,y_t,\a_t)$ can be dropped.
The update rule in (\ref{eq:our_up_x1})
then reduces to MISO  using the Lipschitz gradient surrogate; and
(\ref{eq:omega}) corresponds to the proximal gradient surrogate
\cite{Mairal2013}. SAG, on the other hand, does not have the proximal term $\|x-\xt\|^2$ in its
update rule, and
also cannot handle a nonsmooth $\O$.
When $\psi\ne0$,  (\ref{eq:r}) can be regarded as a quadratic surrogate
\cite{Mairal2013}.
Then,  (\ref{eq:uzawa1}) (resp. (\ref{eq:omega}))
is a combination of the
Lipschitz
(resp.  proximal)
gradient surrogate
and
quadratic surrogate, which can be easily seen to be another surrogate function in the sense of
\cite{Mairal2013}.  However, one should be reminded that MISO only considers unconstrained
optimization
problems and cannot handle the equality constraints in the ADMM setting
(i.e., when $\psi \neq 0$).

The stochastic algorithms in Section~\ref{sec:admm} only
require $\phi$ to be convex,  and
do not explicitly consider
its form in (\ref{eq:phi}). Hence,
there are two possibilities in the handling of a
nonzero $\O$.
The first approach directly takes the nonsmooth $\phi$ in (\ref{eq:phi}), and
uses its subgradient in the update equations ((\ref{eq:ouyang_up_x2}), (\ref{eq:suzuki}) and
(\ref{eq:rda_up_x})). However,  unlike the proximal step in (\ref{eq:our_up_x2}),
this does not exploit the structure of $\phi$ and subgradient descent often has slow empirical
convergence
\cite{Duchi2009}. The second approach
folds $\O$ into $\psi$ by rewriting the optimization problem as
$\min_{x,{\tiny
\begin{bmatrix}
y\\
z
\end{bmatrix}}}
\; \frac 1n\sumn \ell_i(x) + \left[\psi(y)+\O(z)\right]\;:\;
\begin{bmatrix}
A\\
I
\end{bmatrix}
x+
\begin{bmatrix}
B& 0\\
0&-I
\end{bmatrix}
\begin{bmatrix}
y\\
z
\end{bmatrix}
=
\begin{bmatrix}
c\\
0
\end{bmatrix}$.
In the update step for
$\begin{bmatrix}
y\\
z
\end{bmatrix}$,
it is easy to see
from (\ref{eq:up_y})
that $y$ and $z$ are
decoupled and thus can be optimized separately.
In comparison with (\ref{eq:our_up_x2}),
a disadvantage of
this reformulation is that  an
additional variable $z$ (which is of the same size as $x$) has to be introduced. Hence,
it is more computationally expensive
empirically.
Moreover,
the radius of the parameter space
is also increased, leading to bigger
constants in the big-Oh notation of the convergence rate
\cite{Ouyang2013,Suzuki2013}.


\subsection{Convergence Analysis}
\label{sec:conv}

The proposed ADMM algorithm has comparable per-iteration complexity
as the existing stochastic versions in Section~\ref{sec:admm}.
In this section, we show that it also
has a much faster convergence rate.
In the standard convergence analysis of ADMM,
equation (\ref{eq:up_x}) is used for updating
$x$
\cite{He2011,Wang2012}.
In the proposed algorithm,
the loss and feasibility violation  are linearized, making
the analysis more difficult.
Moreover, though related to
MISO,
our analysis is a non-trivial extension
because of the presence of
equality constraints and additional Lagrangian multipliers
in the ADMM formulation.



Let $\|x\|_H\equiv x^THx$ for  a psd matrix $H$,
$H_x\equiv L_AI-\r A^TA$, and $H_y\equiv \r B^TB$.
Moreover, denote the optimal solution of (\ref{eq:obj}) by $(x^*,y^*)$.
As in \cite{Ouyang2013}, we first establish convergence rates of the
$(\bar x_T,\bar y_T)$ solution
in terms of
a combination of the objective value and
feasibility violation (weighted by
$\g>0$).
Proof is in Appendix~\ref{app:cvxLADMM}.

\begin{theorem} \label{thm:cvxLADMM}
Using update rule (\ref{eq:our_up_x2}),
we have
$\E\left[\Phi(\bar x_T,\bar y_T)-\Phi(x^*,y^*)+ \g\|A\bar x_T+B\bar y_T-c\|\right]
\le \frac{1}{2T} \big\{\|x^*-x_0\|_{H_x}^2+nL\|x^*-x_0\|^2 +\|y^*-y_0\|_{H_y}^2  +
{2\r}\left(\frac{\g^2}{\r^2}+\|\a_0\|^2 \right) \big\}$.
\end{theorem}

\begin{remark}
Obviously, (\ref{eq:iu}) reduces to
(\ref{eq:uzawa1}) when $\O=0$. Hence,
Theorem~\ref{thm:cvxLADMM}
trivially holds when the update rule
(\ref{eq:uzawa}) is used.
\end{remark}

When $\O=0$ and the inexact Uzawa simplification is not used,
a similar convergence rate can be obtained  in the following.
Proof is in Appendix~\ref{app:cvxLADMM2}.

\begin{theorem} \label{thm:cvxLADMM2}
Using update rule (\ref{eq:our_upd_x}), we have
$\E\left[\Phi(\bar x_T,\bar y_T)-\Phi(x^*,y^*)+ \g\|A\bar x_T+B\bar y_T-c\|\right]
\le \frac{1}{2T} \left\{nL\|x^*-x_0\|^2 +\|y^*-y_0\|_{H_y}^2  +
{2\r}\left(\frac{\g^2}{\r^2}+\|\a_0\|^2 \right) \right\}$.
\end{theorem}

As in other ADMM algorithms,
the $(\bar x_T,\bar y_T)$ pair obtained from
Algorithm~\ref{alg:sadmm}
may not
satisfy the linear constraint
$Ax+By=c$ exactly. As discussed in \cite{Suzuki2013},
when $B$ is invertible, this can be alleviated by obtaining $y$
from $\bar x_T$
as $y(\bar x_T)=B^{-1}(c-A\bar x_T)$.
The feasibility violation is then zero, and
the following corollary shows convergence w.r.t. the objective value.
Proof is in Appendix~\ref{app:cor}.
Similarly,
when $A$ is invertible.
one can also obtain $x$
from $\bar y_T$
as $x(\bar y_T)=A^{-1}(c-B\bar y_T)$. Because of the lack of space, the analogous corollary will
not be shown here.

\begin{corollary}\label{cor:cvxLADMM}
Assume that
$\psi$ is $\tL$-Lipschitz continuous, and
$B$ is invertible.
Using the update rule (\ref{eq:uzawa}) or (\ref{eq:our_up_x2}), we have
$\E\left[\Phi(\bar x_T,y(\bar x_T)-\Phi(x^*,y^*)\right]
\le \frac{1}{2T} \big\{\|x^*-x_0\|_{H_x}^2+nL\|x^*-x_0\|^2+\|y^*-y_0\|_{H_y}^2  +
{\r}\left(\frac{\tL^2{L_B}}{\r^2}+\|\a_0\|^2\right) \big\}$,
where $L_B$ is the largest eigenvalue of $(B^{-1})^TB^{-1}$;
when update rule (\ref{eq:our_upd_x}) is used,
$\E\left[\Phi(\bar x_T,y(\bar x_T)-\Phi(x^*,y^*)\right]
\le\frac{1}{2T} \left\{nL\|x^*-x_0\|^2+\|y^*-y_0\|_{H_y}^2  +
{\r}\left(\frac{\tL^2{L_B}}{\r^2}+\|\a_0\|^2\right) \right\}$.
\end{corollary}

\begin{remark}
In all the above cases, we obtain a convergence rate of
$\mO(1/T)$,
which matches that of the batch ADMM but with a much lower per-iteration complexity.
\end{remark}




\section{Experiments}
\label{sec:expt}

In this section, we perform experiments on
the generalized lasso model \cite{Tibshirani2011a}:
$\min_{x\in \mathbb{R}^d} \; \frac 1 {n} \sum_{i=1}^n \ell_i(x)
+ \l\|Ax\|_1$,
where
$A\in \mathbb{R}^{m\times d}$
(for some $m>0$)
is a penalty matrix specifying the desired structured sparsity pattern of $x$. With different settings
of $A$, this can be reduced to  models such as
the fused lasso, trend filtering, and wavelet smoothing.
Here, we will focus on the graph-guided fused lasso \cite{kim-09},
whose sparsity pattern
is specified by a graph defined
on the $d$ variates of $x$.
Following \cite{Ouyang2013}, we obtain this graph by
sparse inverse covariance selection \cite{Banerjee2008, Boyd2010}.
Moreover, as
classification problems
are considered here, we
use the logistic loss
instead of the square loss commonly used in lasso.

\begin{figure}[t]
\begin{center}
\begin{tabular}{c}
\subfigure[a9a]{\includegraphics[height=1.3in]{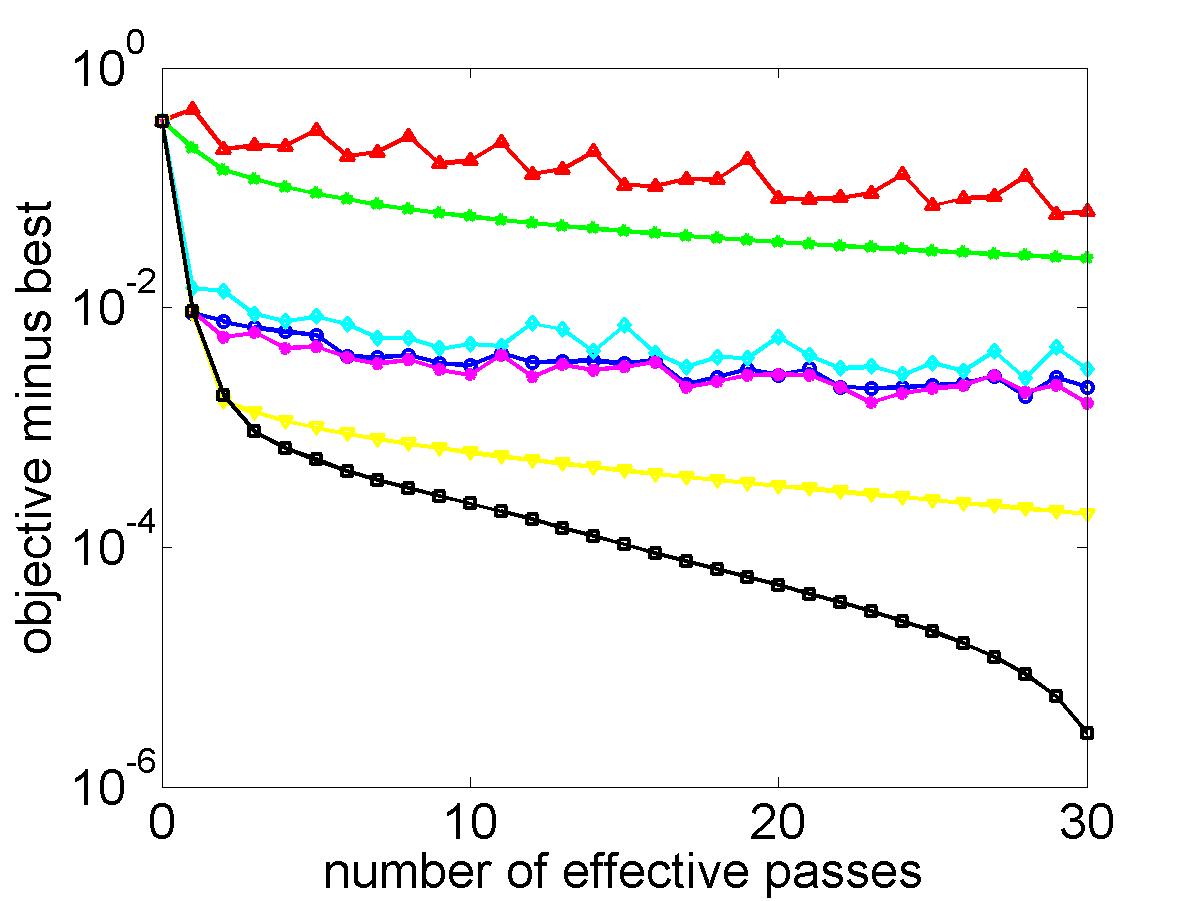}}
\subfigure[covertype]{\includegraphics[height=1.3in]{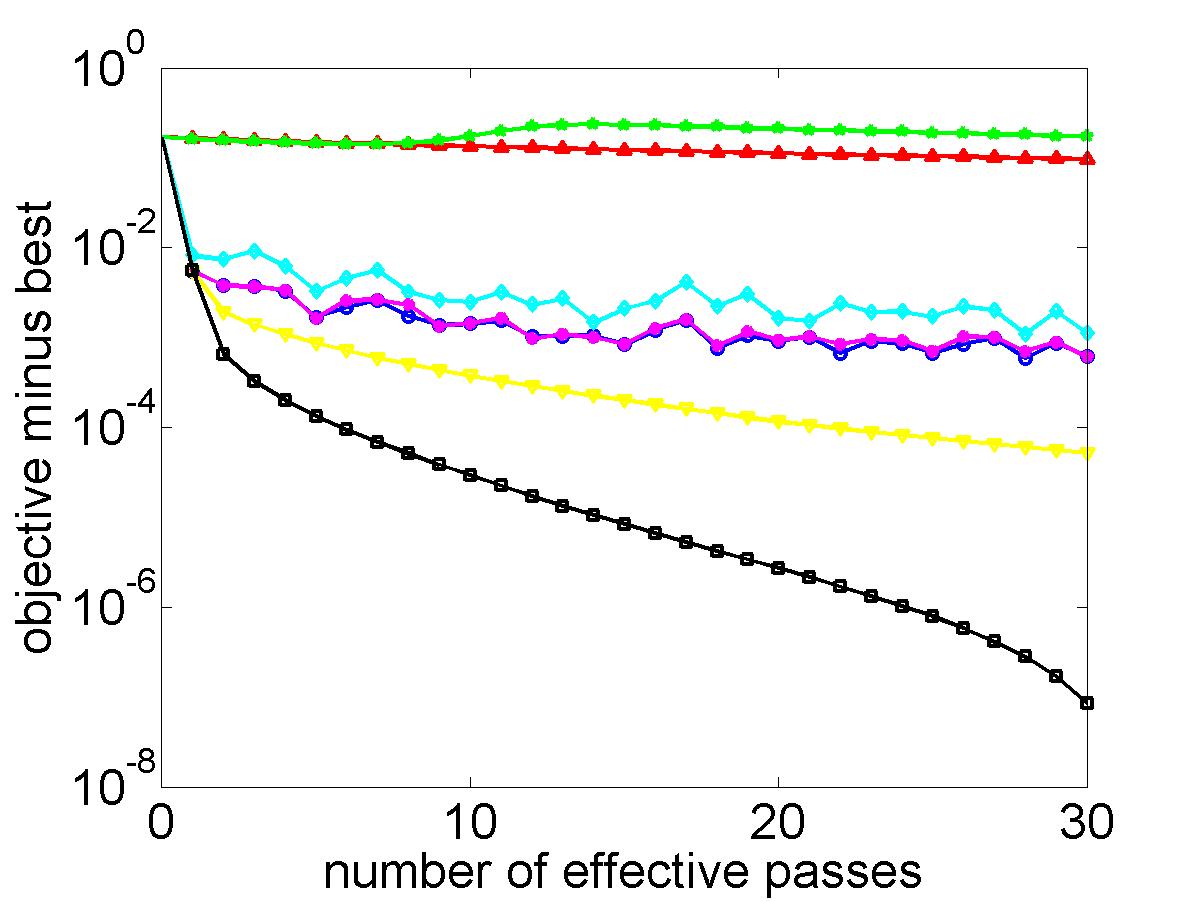}}\\
\subfigure[quantum]{\includegraphics[height=1.3in]{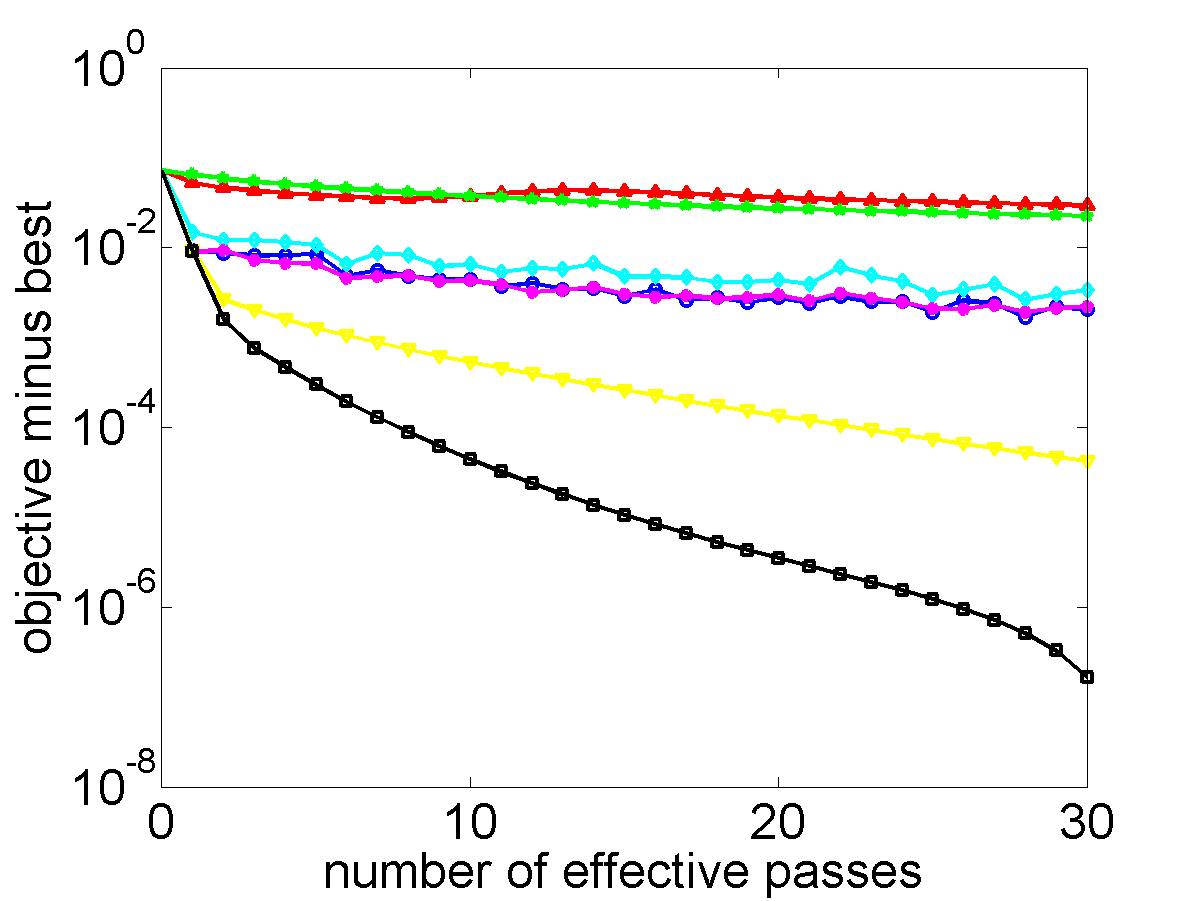}}
\subfigure[rcv1]{\includegraphics[height=1.3in]{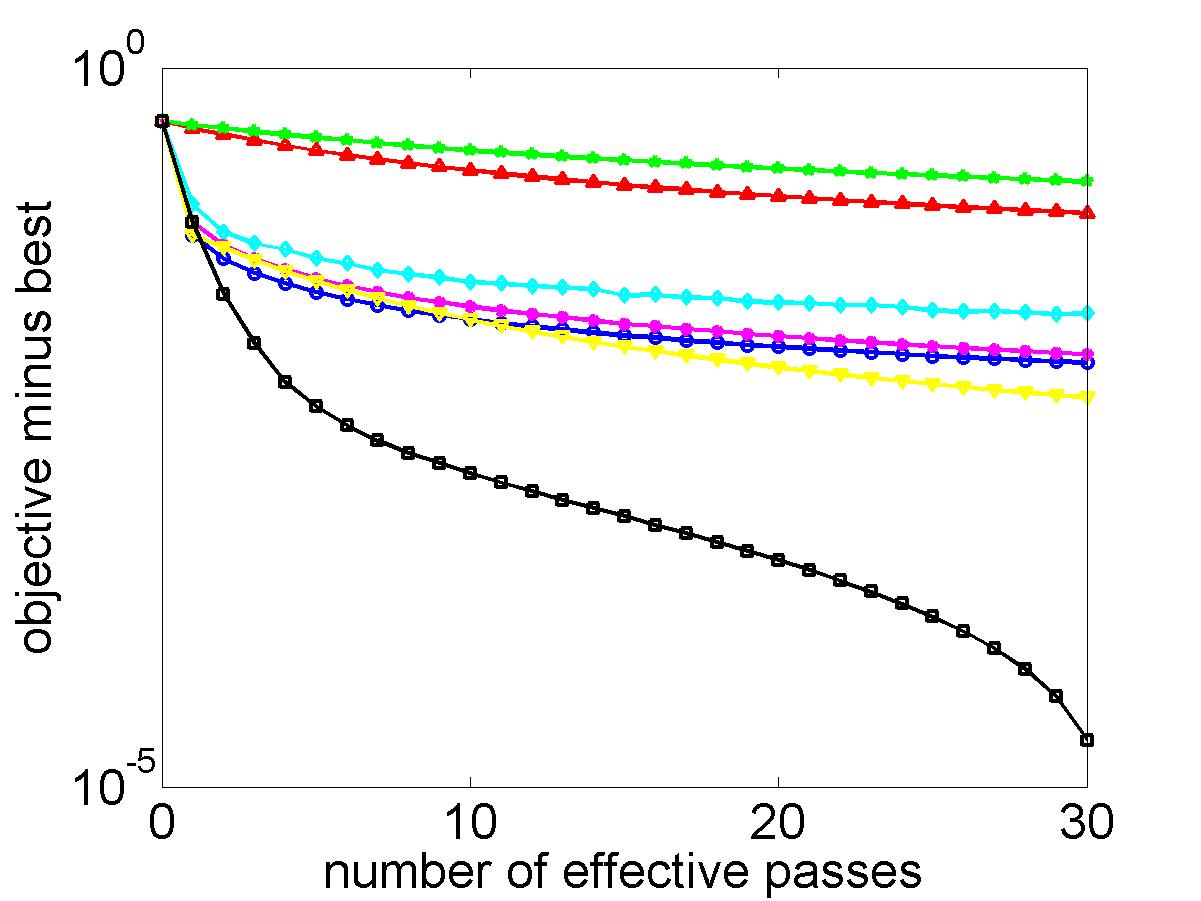}}\\
\subfigure[sido]{\includegraphics[height=1.3in]{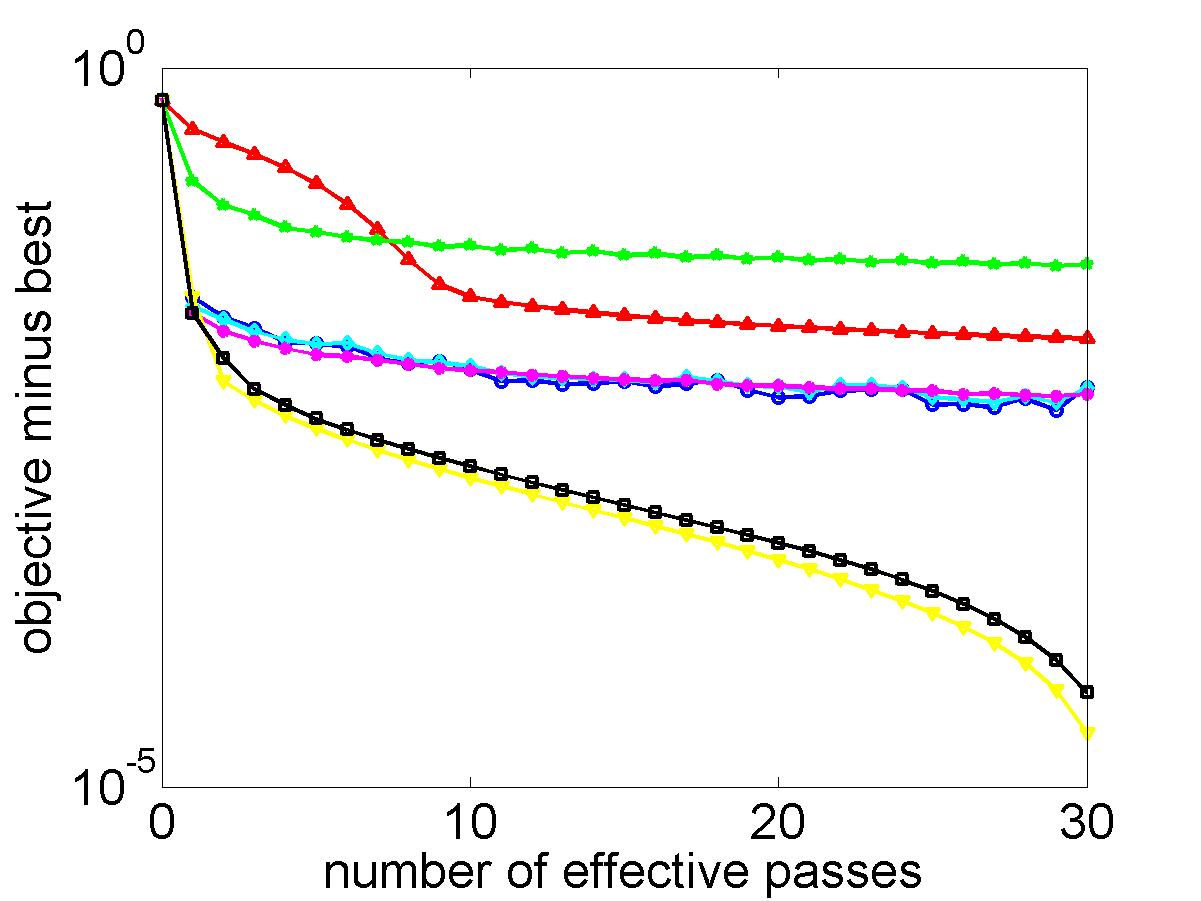}} \\
\subfigure{\includegraphics[width=.45\textwidth]{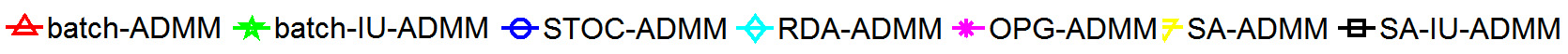}}
\end{tabular}
\end{center}
\vspace{-.2in}
\caption{Objective value obtained
versus number of effective passes over data.
In the plots,
the ``best'' objective value
is defined
as the lowest value obtained across the methods.}
\label{fig:convex_obj}
\end{figure}

While proximal methods
have been used in the
optimization of
graph-guided fused lasso
\cite{Barbero2011,Liu2010},
in general,
the underlying proximal step does not have a closed-form
because of the presence of $A$.

ADMM,
by
splitting the objective as
$\phi(x)
=\frac 1 {n} \sum_{i=1}^n \ell_i(x),
\psi(y)=\l\|y\|_1$ and with constraint $Ax=y$,
has been shown to be more
efficient
\cite{Ouyang2013,Suzuki2013}.
In this experiment, we compare
\begin{enumerate}
\item two variants  of the proposed method: i) SA-ADMM, which uses update rule
(\ref{eq:our_upd_x}); and
ii) SA-IU-ADMM, which uses (\ref{eq:uzawa})) based on the inexact Uzawa method;
\item three  existing
stochastic ADMM algorithms:
i) STOC-ADMM~\cite{Ouyang2013};
ii) OPG-ADMM~\cite{Suzuki2013};  and
iii) RDA-ADMM \cite{Suzuki2013};
\item two deterministic
ADMM variants:
i) batch-ADMM, which is the batch version of SA-ADMM using
full gradient
(i.e., $\tau_i(t)=t \; \forall i$);
and ii)
batch-IU-ADMM, which is
the batch version of SA-IU-ADMM.
\end{enumerate}
All these share the same update rules for $y$ and $\alpha$ (i.e., steps~\ref{alg:sadmm_upy} and \ref{alg:sadmm_upa} in Algorithm~\ref{alg:sadmm}), and differ only
in the updating
of $x$, which are summarized in Table~\ref{tbl:algs}.
We do not compare with the online ADMM~\cite{Wang2012} or
a direct application of batch ADMM, as they
require nonlinear optimization for the update of $x$.
Moreover,  it has been
shown that
the online ADMM
is slower than RDA-ADMM \cite{Suzuki2013}.

\begin{table}[htbp]
\caption{A comparison of the update rules on $x$.}
\label{tbl:algs}
{\footnotesize
\begin{tabular}{c|c|c|c} \hline
&gradient &linearize& constant\\
&computation &$\scriptstyle \|Ax+By-c\|^2$? &stepsize?\\ \hline
SA-ADMM& average &no& yes\\
SA-IU-ADMM& average &yes& yes\\
STOC-ADMM & one sample&no& no \\
OPG-ADMM &one sample&yes& no \\
RDA-ADMM &average (history) &yes& no\\
batch-ADMM&all samples & no& yes\\
batch-IU-ADMM&all samples&yes& yes\\ \hline
\end{tabular}}
\end{table}

\begin{table}[htbp]
\caption{Summary of data sets.}
\label{tbl:data}
\begin{center}
{\footnotesize
\begin{tabular}{c|c|c}
\hline data set&\#samples &dim  \\ \hline
a9a&32,561&123 \\
covertype&581,012&54\\
quantum&50,000&78\\
rcv1&20,242&47,236\\
sido&12,678&4,932\\
\hline
\end{tabular}}
\end{center}
\hfill
\end{table}


Experiments are performed on five binary classification data sets:\footnote{{\em a9a}, {\em
covertype} and  {\em rcv1} are from the LIBSVM archive,
{\em quantum} is from the KDDCup-2004,
and {\em sido} from the Causality Workbench website.}
{\em a9a},
{\em covertype}, {\em quantum},
{\em rcv1},
and
{\em sido}
(Table~\ref{tbl:data}), which have been commonly used \cite{LeRoux2012,Suzuki2013}.
For each data set, half of the samples are used for training, while the rest for testing.
To reduce statistical variability, results are averaged over 10 repetitions.
We fix
$\rho$
in (\ref{eq:lag})
to $0.01$; and the regularization parameter
$\lambda$ to $10^{-5}$ for {\em a9a, covertype, quantum}, and to
$10^{-4}$ for {\em rcv1} and {\em sido}.
For selection of stepsize (or its proportionality constant),
we run
each stochastic algorithm
5 times over a small training subset (with 500 samples) and choose the setting with the
smallest training objective value. For each batch algorithm,
we run it for 100 iterations
on the same training subset.
All methods are
implemented in MATLAB.

\begin{figure}[htbp]
\begin{center}
\begin{tabular}{c}
\subfigure[a9a]{\includegraphics[height=1.3in]{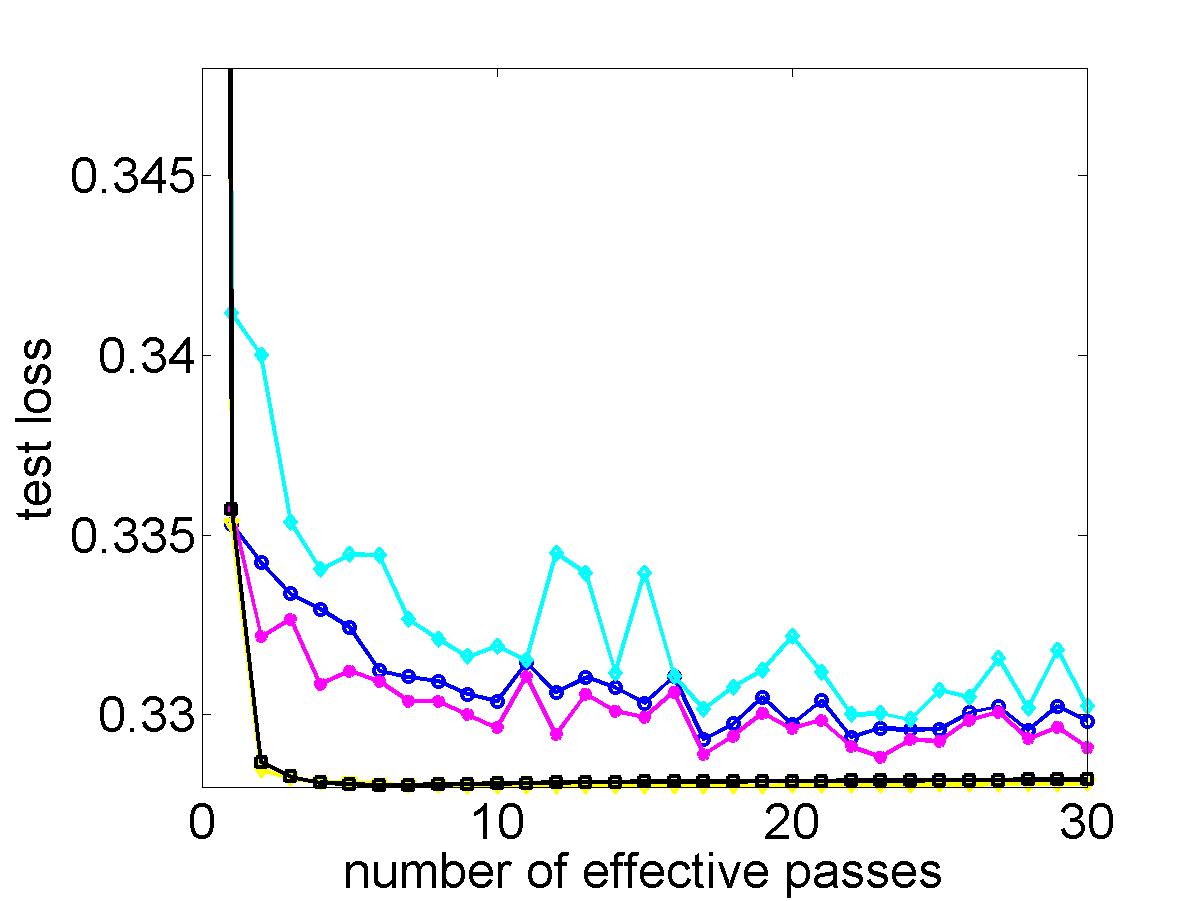}}
\subfigure[covertype]{\includegraphics[height=1.3in]{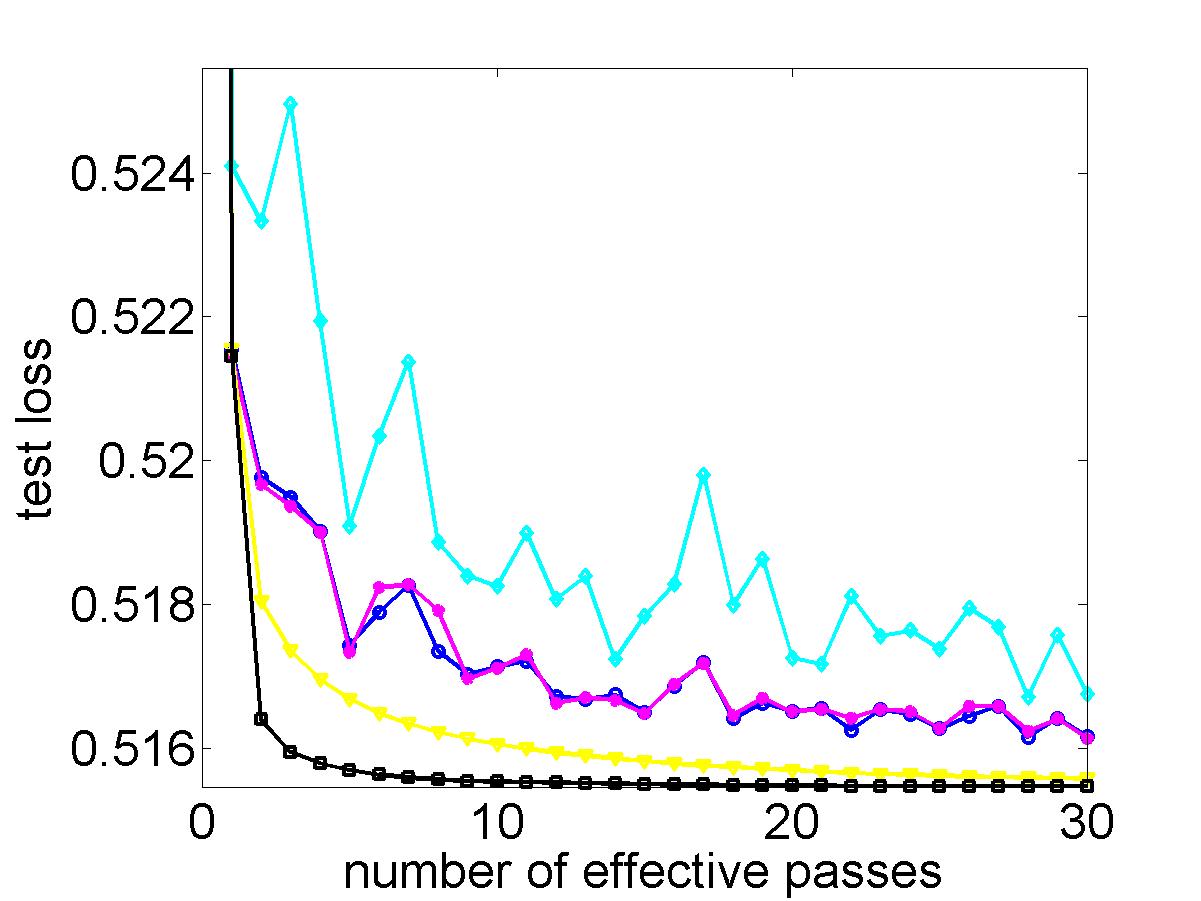}}\\
\subfigure[quantum]{\includegraphics[height=1.3in]{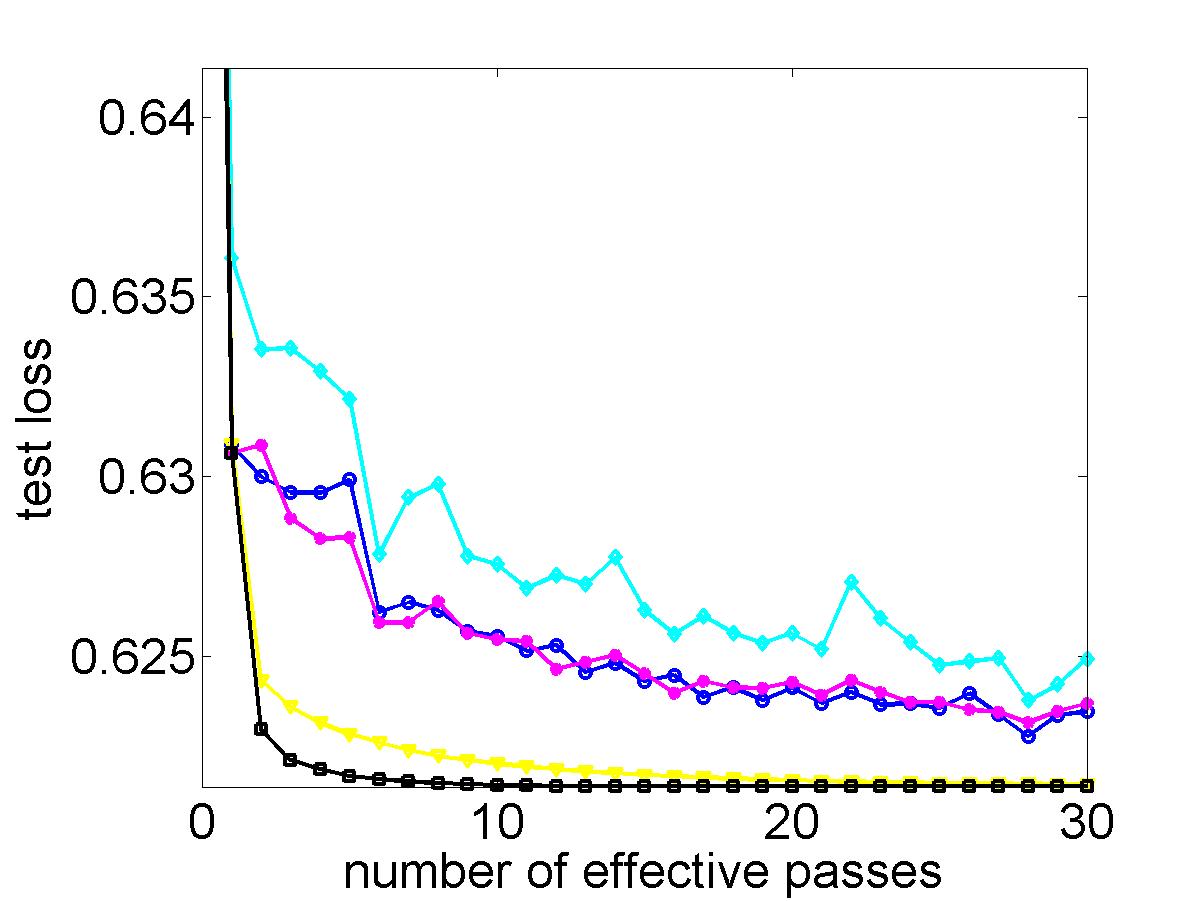}}
\subfigure[rcv1]{\includegraphics[height=1.3in]{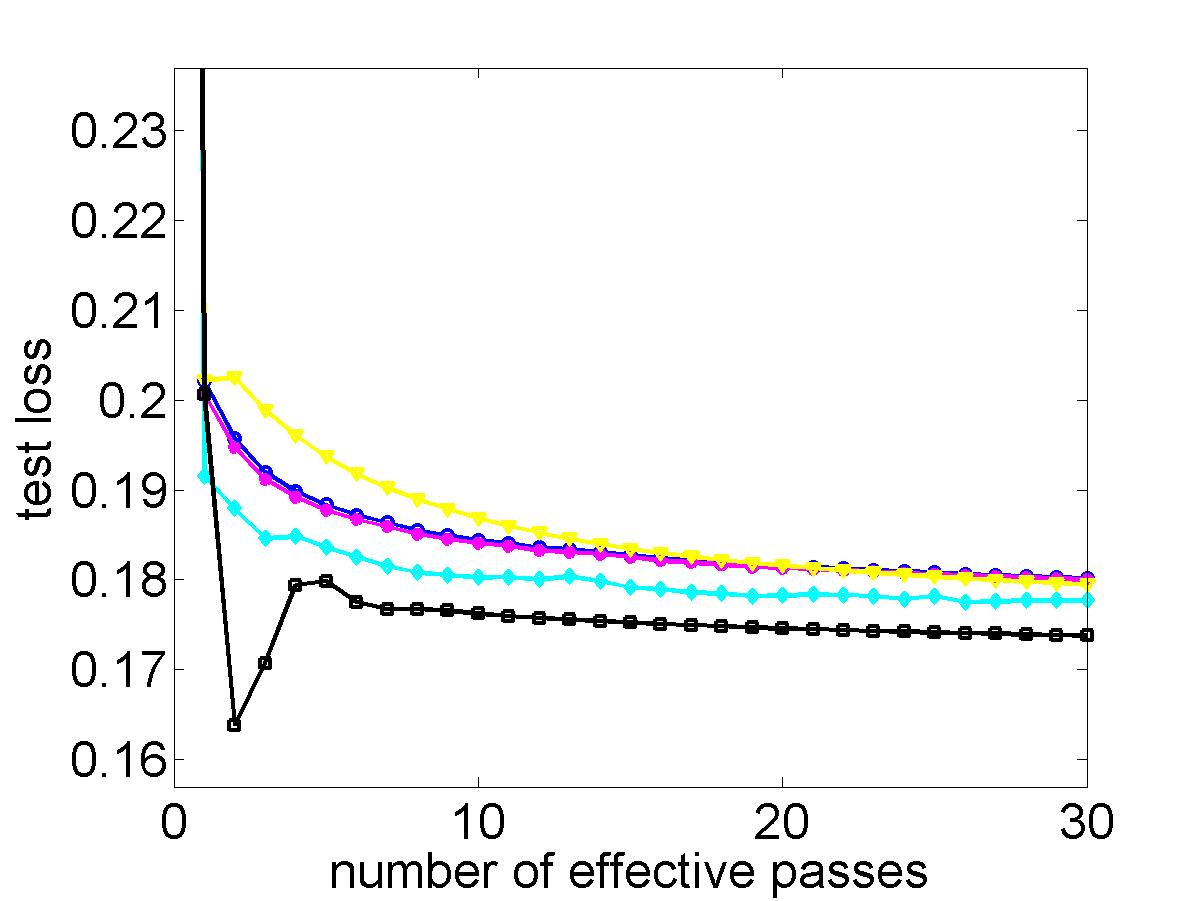}}\\
\subfigure[sido]{\includegraphics[height=1.3in]{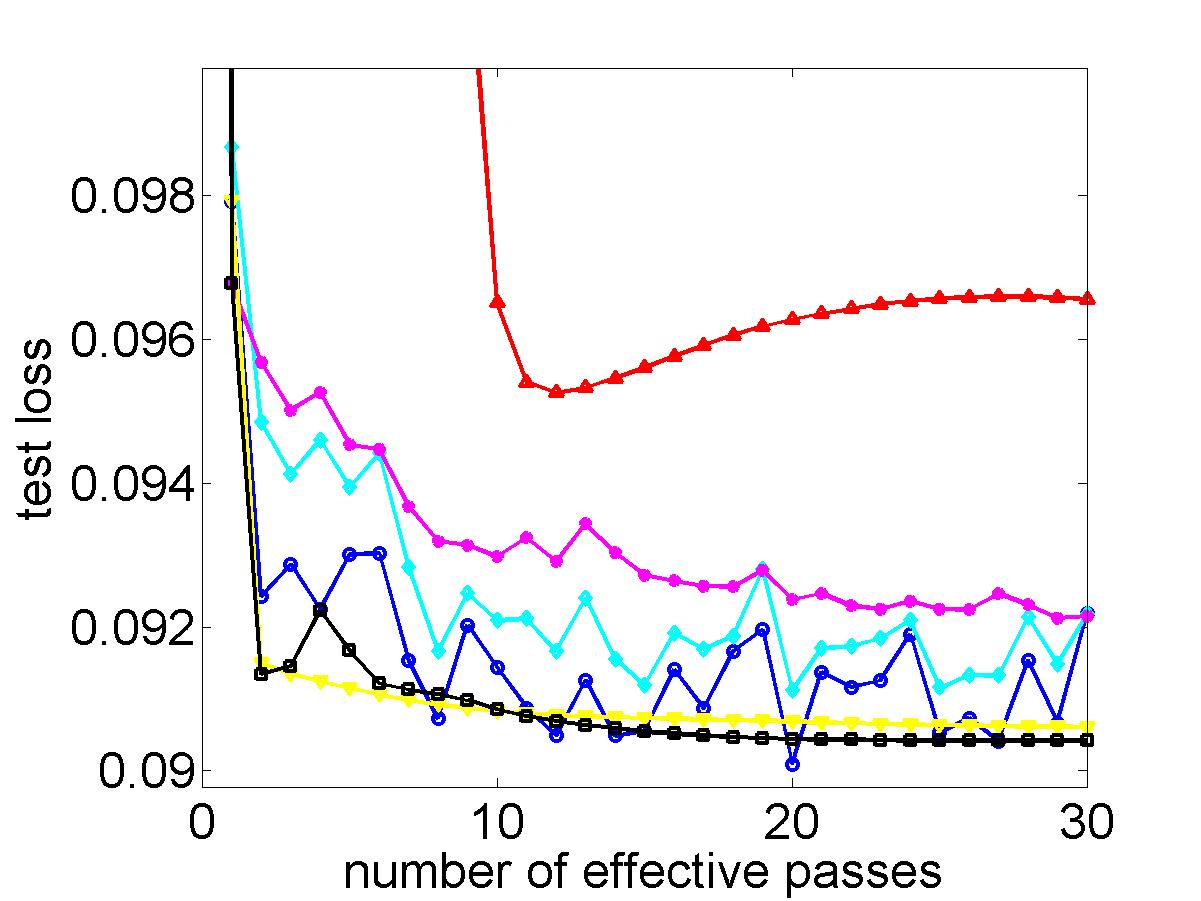}}\\
\subfigure{\includegraphics[width=.45\textwidth]{legend.jpg}}
\end{tabular}
\end{center}
\vspace{-.2in}
\caption{Testing loss versus number of effective passes.}
\label{fig:convex_pass_loss}
\end{figure}

Figures~\ref{fig:convex_obj} and \ref{fig:convex_pass_loss} show
 the objective value and
testing loss
versus the number of
effective passes over the data.
Overall, SA-IU-ADMM is the fastest, which is followed by SA-ADMM,
and then the other stochastic algorithms. The batch ADMM algorithms are the slowest.

\begin{figure}[htbp]
\begin{center}
\begin{tabular}{c}
\subfigure[a9a]{\includegraphics[height=1.3in]{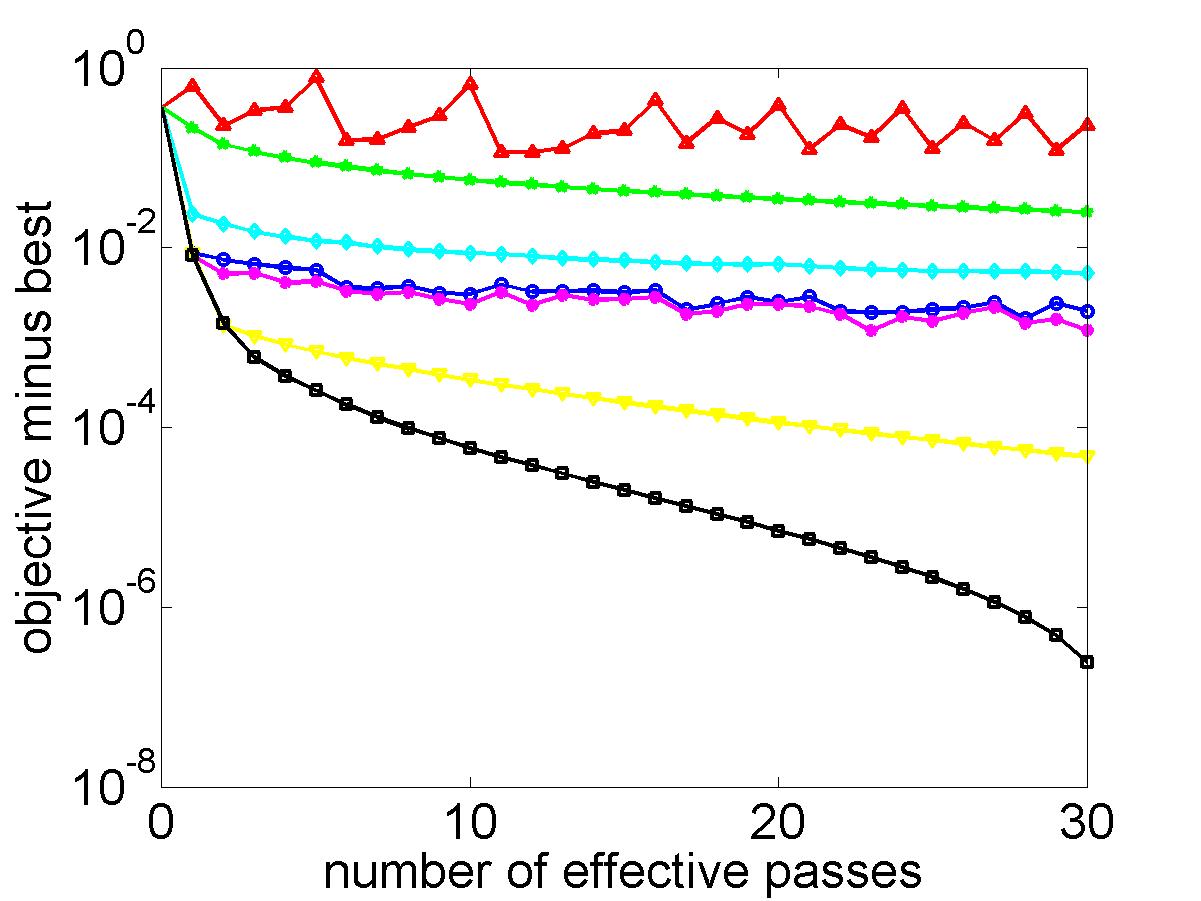}}
\subfigure[covertype]{\includegraphics[height=1.3in]{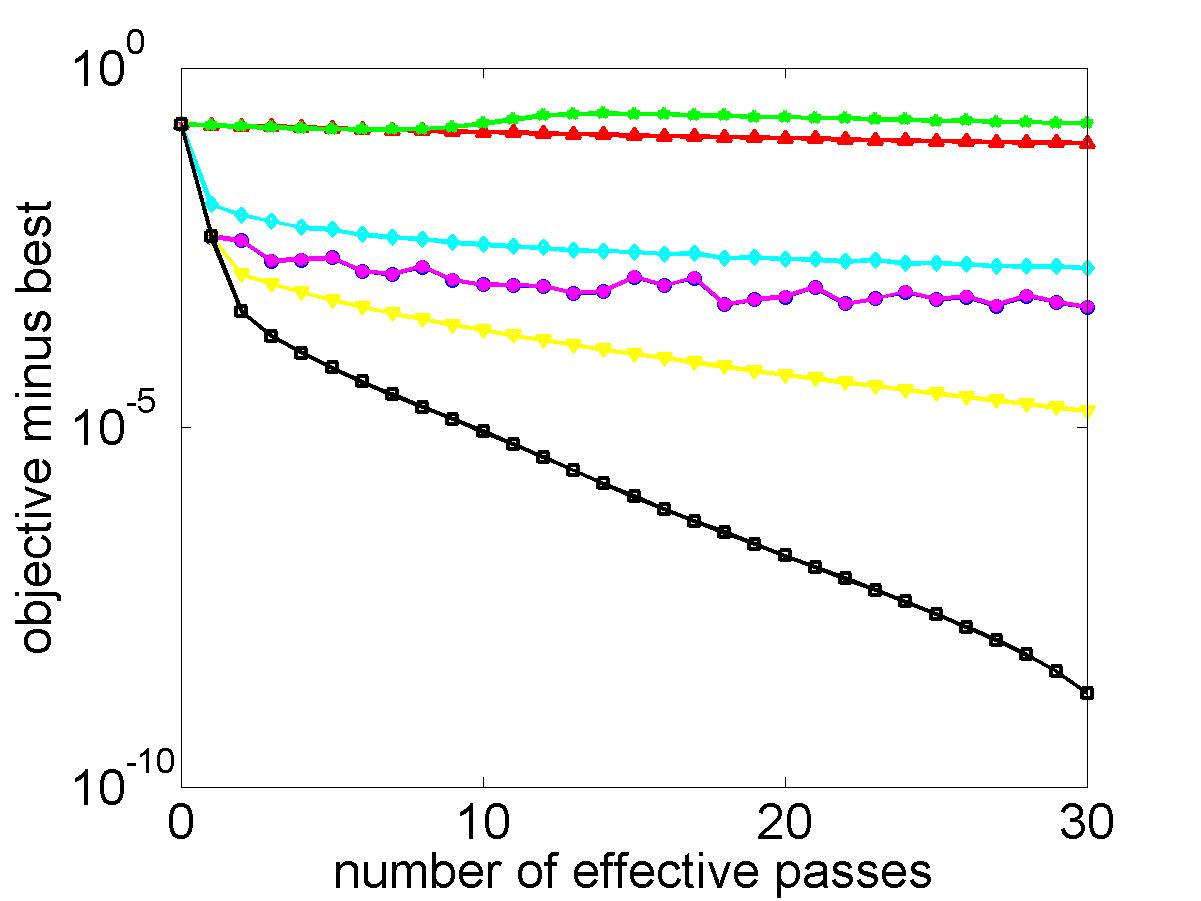}}\\
\subfigure[quantum]{\includegraphics[height=1.3in]{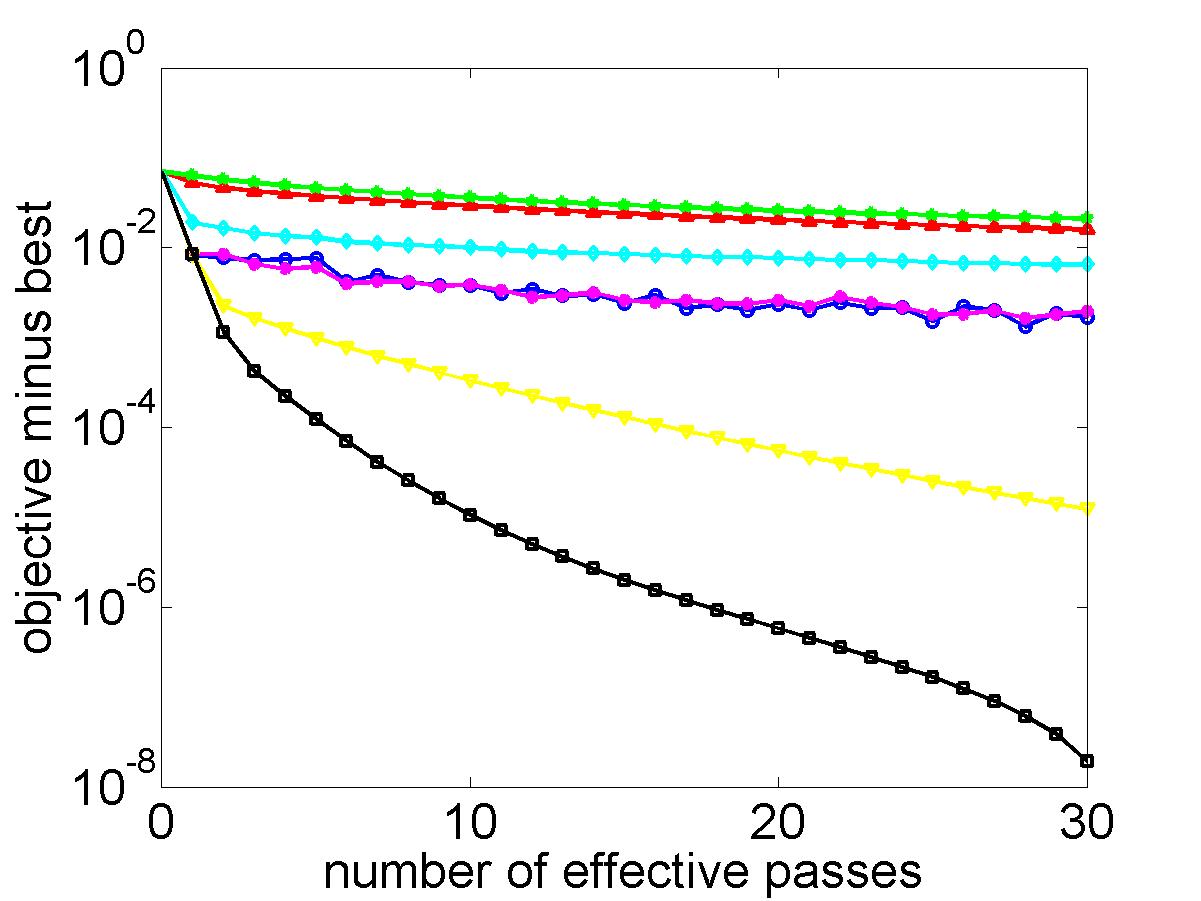}}
\subfigure[rcv1]{\includegraphics[height=1.3in]{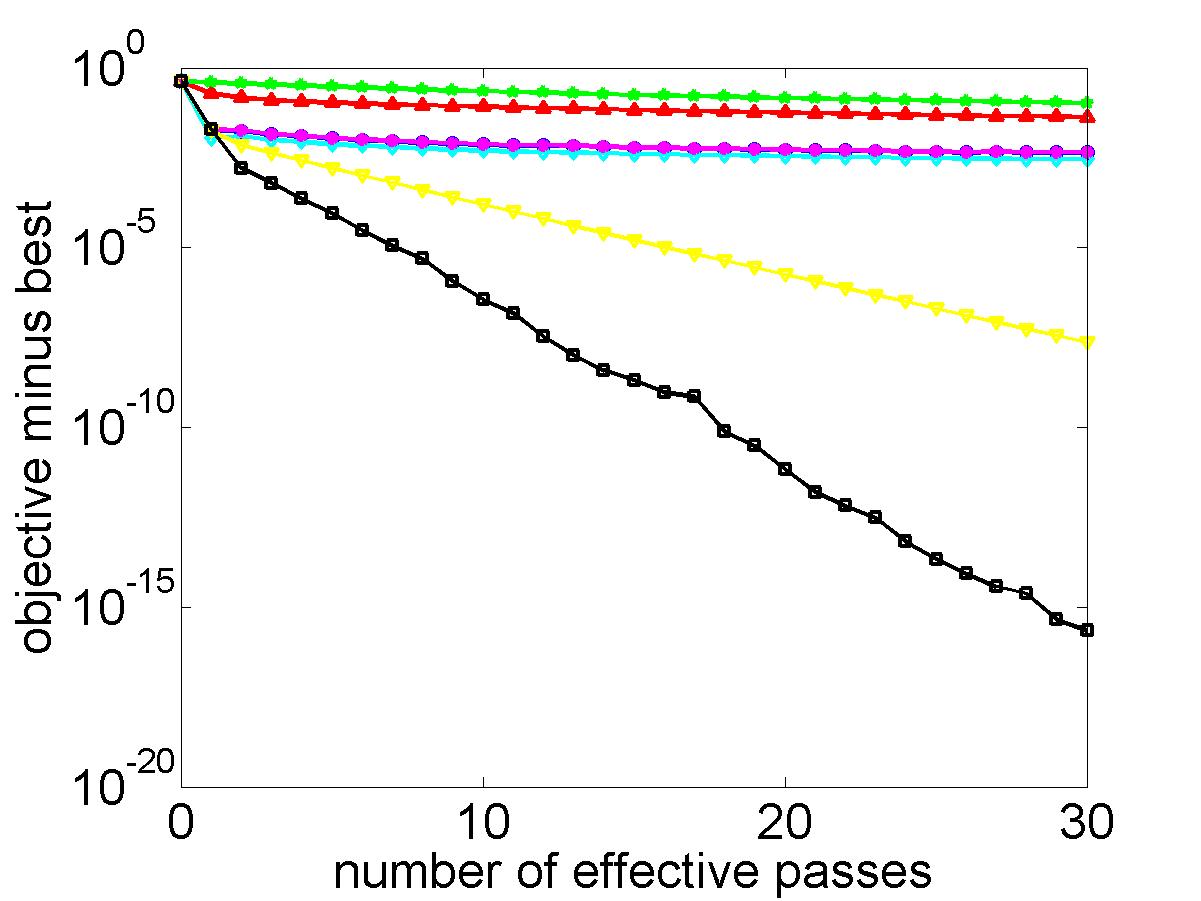}}\\
\subfigure[sido]{\includegraphics[height=1.3in]{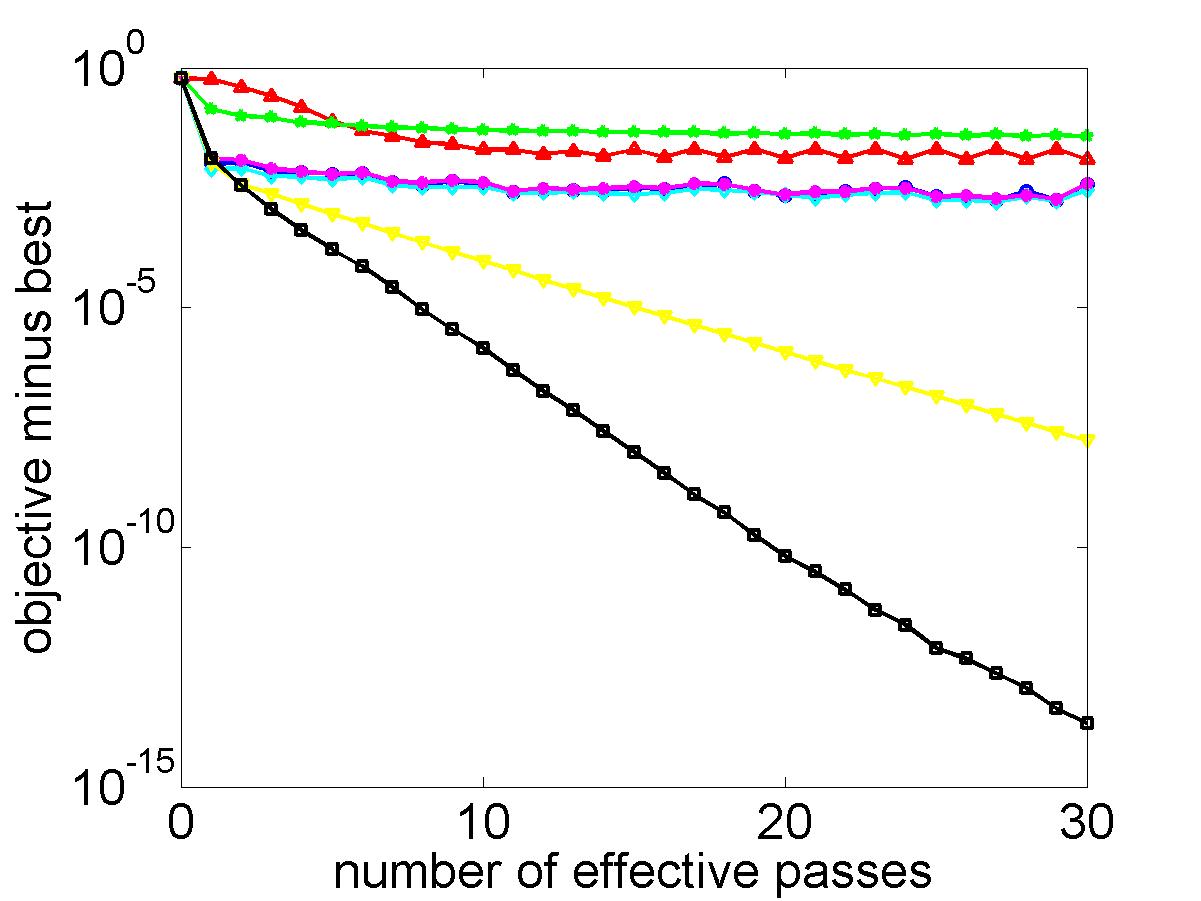}}\\
\\
\subfigure{\includegraphics[width=.45\textwidth]{legend.jpg}}
\end{tabular}
\end{center}
\vspace{-.2in}
\caption{Objective value versus number of effective passes for the strongly convex problem.}
\label{fig:str_convex_obj}
\end{figure}

As discussed in Section~\ref{sec:admm},
STOC-ADMM and OPG-ADMM
has $\mO(\log T/T)$ convergence when the loss is strongly
convex.
It is still an open question
whether this also holds for the proposed algorithm.
In the following,
we will compare their performance empirically
by adding an extra $\ell_2$-regularizer on $x$.
Results
are shown in Figure~\ref{fig:str_convex_obj}.
As can be seen, the
improvement of SA-IU-ADMM over others is even more dramatic.

\section{Conclusion}

In this paper, we developed a novel stochastic algorithm that
incrementally approximates the full gradient in the linearized ADMM formulation.
It enjoys the same computational simplicity  as existing
stochastic ADMM algorithms, but has a fast
convergence rate that matches the batch ADMM. Empirical results on both general
convex and strongly convex problems demonstrate its efficiency over
batch and stochastic ADMM algorithms.
In the
future, we will investigate the theoretical convergence rate of the proposed algorithm
on strongly convex problems.

\bibliographystyle{IEEEtran}
\bibliography{MyReference}

\end{document}